\documentclass[letterpaper, 10pt, conference]{IEEEtran}
\IEEEoverridecommandlockouts

\usepackage{amsmath}
\usepackage{amssymb}
\usepackage{graphicx}
\usepackage{booktabs}
\usepackage{multirow}
\usepackage{xcolor}
\usepackage{cite}
\usepackage{url}
\usepackage{placeins}
\usepackage{dblfloatfix}
\usepackage{capt-of}
\usepackage[hidelinks]{hyperref}
\DeclareMathSizes{6.3}{6.3}{5}{5}

\title{\LARGE\bfseries
StableMimic: Smooth Human-Like Recovery\\[-0.05em]
for Humanoid Motion Tracking\\[0.25em]
{\normalsize\mdseries\itshape Learning Beyond the Tracking Distribution for Structured Post-Fall Behavior}
}

\author{
Weihao Wu, Ming Huang, Ruofei Liu, Jinglei Nie, Shuxiang Guo$^{*}$, and Chunying Li$^{*}$%
\thanks{All authors are with the Southern University of Science and Technology (SUSTech), Shenzhen, China.}%
\thanks{$^{*}$Corresponding authors: Shuxiang Guo and Chunying Li.}
}

\newcommand{\cmark}{\textcolor{green!50!black}{\checkmark}}
\newcommand{\xmark}{\textcolor{red!75!black}{\ensuremath{\times}}}
\newcommand{\TrackingMlpMPBPE}{32.66}
\newcommand{\TrackingMoeMPBPE}{28.53}
\newcommand{\TrackingMlpMJAE}{113.56}
\newcommand{\TrackingMoeMJAE}{88.83}
\newcommand{\RecoveryMlpSuccess}{98/100}
\newcommand{\RecoveryMoeSuccess}{100/100}
\newcommand{\RecoveryMlpLimbSpeedTail}{6.06}
\newcommand{\RecoveryMoeLimbSpeedTail}{4.91}
\newcommand{\RecoveryMlpLimbPath}{16.04}
\newcommand{\RecoveryMoeLimbPath}{13.80}
\newcommand{\RecoveryMlpJointSpeed}{6.34}
\newcommand{\RecoveryMoeJointSpeed}{5.26}
\newcommand{\RecoveryMlpTorqueTail}{27.94}
\newcommand{\RecoveryMoeTorqueTail}{19.64}
\newcommand{\RecoveryMlpEnergy}{1.10}
\newcommand{\RecoveryMoeEnergy}{0.61}

\begin{document}

\IEEEaftertitletext{%
  \begin{minipage}{\textwidth}
    \centering
    \includegraphics[width=\textwidth]{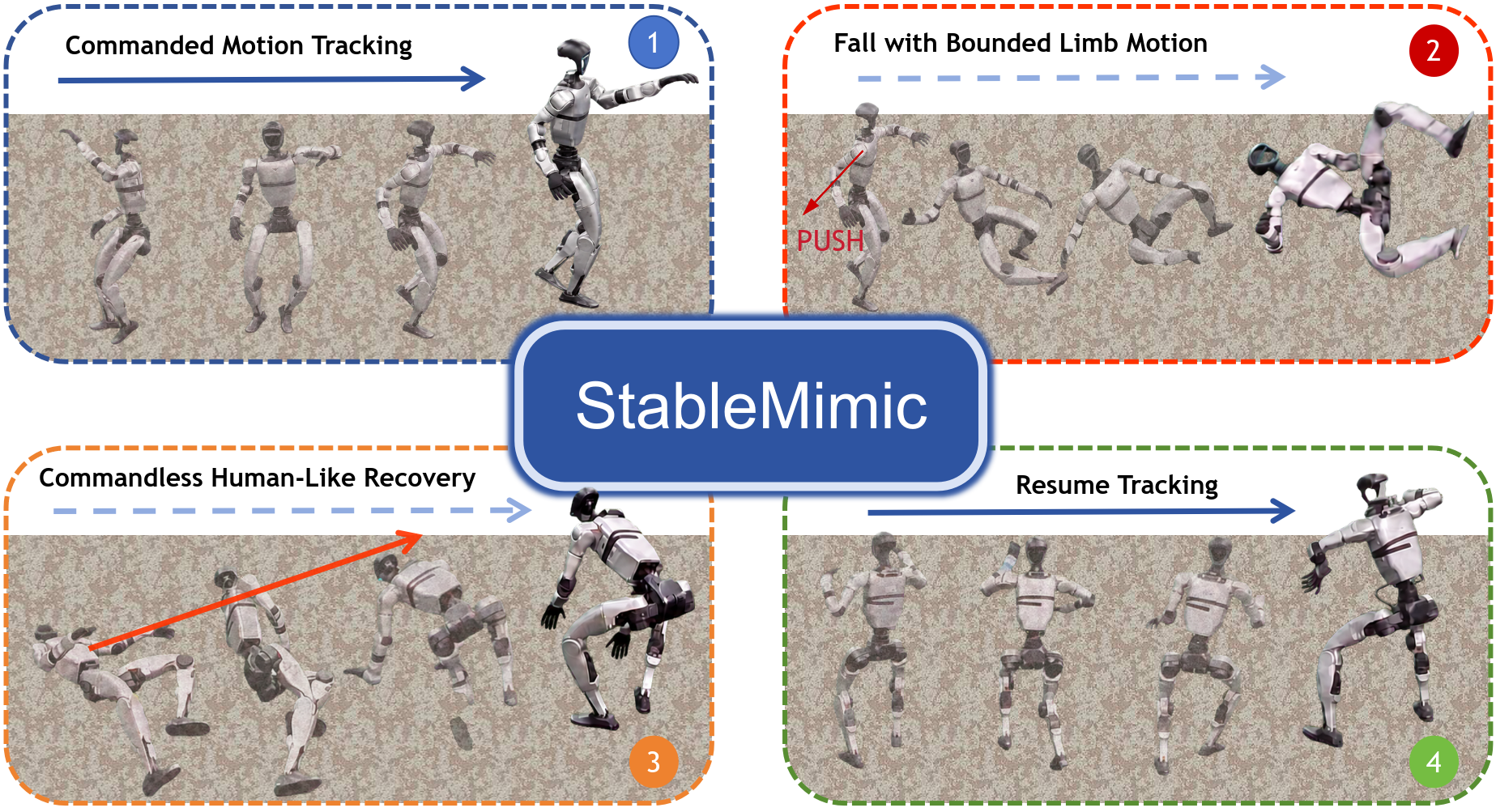}
    \refstepcounter{figure}\label{fig:commandless_recovery_overview}

    \vspace{2pt}
    {\footnotesize Fig.~\thefigure.\enspace StableMimic overview. One policy tracks a continuously advancing whole-body command, routes fall-induced proprioceptive states toward recovery, and resumes tracking after command reacquisition. No external recovery trigger or runtime get-up reference is used.\par}
  \end{minipage}
  \vspace{8pt}
}

\maketitle
\thispagestyle{empty}
\pagestyle{empty}

\begin{abstract}
Humanoid motion trackers perform reliably within learned tracking distributions, but falls can move the robot into low-height, contact-rich states from which an advancing command is temporarily unreachable.
Tracking-only policies may chase infeasible references, producing rapid, large-amplitude limb corrections that increase risk to the robot and its surroundings.
We present StableMimic, a unified tracker trained beyond the nominal tracking distribution.
Perturbed resets around multiple human get-up references expose prone, supine, off-balance, and intermediate ground-contact states, shaping structured recovery that returns the robot to the trackable region.
Because tracking and recovery occupy markedly different state--action distributions, StableMimic uses dedicated experts for each regime and a proprioceptive gate that continuously blends their actions.
A hidden successor-state objective teaches human-reference-shaped recovery without exposing reference identity or phase to the deployed Actor; deployment requires no get-up reference, recovery command, trajectory retrieval, or external policy switch.
On the complete retargeted LAFAN1 dance subset, StableMimic achieves the lowest errors on all four tracking metrics among five methods.
Across 100 matched push-to-fall trials per method, it recovers in 100/100 and attains the lowest values on six of seven post-fall motion and load measures, supporting improved interaction safety under this protocol.
Real Unitree G1 dance and standing-reference deployments qualitatively demonstrate bounded limb motion, autonomous recovery, and command resumption.
\end{abstract}

\begin{table*}[!t]
  \centering
  \caption{Comparison of representative humanoid tracking and recovery systems.
  \textcolor{blue!65!black}{Human Get-Up Ref.}: ``Training required'' means training only;
  \textcolor{blue!65!black}{No Get-Up Cmd.}: no separate command or recovery controller;
  \textcolor{blue!65!black}{Training Compute}: qualitative demand of the complete pipeline.
  Checkmarks denote explicit support and ``--'' not applicable.}
  \label{tab:literature_positioning}
  \footnotesize
  \renewcommand{\arraystretch}{1.12}
  \begin{tabular*}{\textwidth}{@{\extracolsep{\fill}}lcccccc@{}}
    \toprule
    \textbf{System} &
    \shortstack{\textbf{Motion}\\\textbf{Tracking}} &
    \shortstack{\textbf{Autonomous}\\\textbf{Recovery}} &
    \shortstack{\textbf{Human Get-Up}\\\textbf{Ref.}} &
    \shortstack{\textbf{No Get-Up}\\\textbf{Cmd.}} &
    \shortstack{\textbf{Resume}\\\textbf{Command}} &
    \shortstack{\textbf{Training}\\\textbf{Compute}} \\
    \midrule
    BeyondMimic~\cite{liao2025beyondmimic} & \cmark & \xmark & \xmark & -- & \xmark & Moderate \\
    SONIC~\cite{luo2025sonic} & \cmark & \xmark & \xmark & -- & \xmark & High \\
    KungFuAthlete~\cite{lei2026kungfuathlete} & \cmark & \cmark & \xmark & \cmark & \cmark & Moderate \\
    BFM-Zero~\cite{li2025bfmzero} & \cmark & \cmark & \textbf{Training required} & \cmark & \cmark & High \\
    HoST~\cite{huang2025host} & \xmark & \cmark & \xmark & \xmark & \xmark & Moderate \\
    HumanUP~\cite{he2025getup} & \xmark & \cmark & \xmark & \xmark & \xmark & Moderate \\
    \midrule
    \textbf{StableMimic (Ours)} & \cmark & \cmark & \textbf{Training required} & \cmark & \cmark & Moderate \\
    \bottomrule
  \end{tabular*}
\end{table*}

\section{Introduction}
\label{sec:introduction}

Humanoid motion trackers support animation, motion capture, and teleoperation, but unexpected contacts and modeling errors can drive them outside the command's trackable distribution, where chasing an infeasible reference produces rapid, large-workspace post-fall motion that increases risk to the robot and its surroundings.
Existing pipelines often separate tracking from stand-up recovery~\cite{gaspard2024frasa,huang2025host,he2025getup}, whereas tracking-centric objectives such as BeyondMimic and SONIC do not prescribe a human-shaped response outside their learned tracking corridor~\cite{liao2025beyondmimic,luo2025sonic}.

StableMimic addresses this gap with one deployable policy for command tracking and human-reference-shaped recovery.
Perturbed get-up resets expand training beyond the command distribution, while hidden successor-state rewards guide recovery and command resumption without exposing reference identity, phase, or a recovery command.

Because tracking and recovery occupy markedly different distributions, a proprioception-gated dual-expert MoE dedicates one expert to each regime and continuously blends their actions without a privileged recovery flag or hard runtime switch.
Fig.~\ref{fig:commandless_recovery_overview} summarizes the tracking--disturbance--recovery--reacquisition lifecycle.

Our contributions are:
\begin{itemize}
  \setlength{\itemsep}{0pt}
  \setlength{\parskip}{0pt}
  \setlength{\parsep}{0pt}
  \item A proprioception-gated dual-expert MoE that blends tracking and recovery within one deployable policy.
  \item A lightweight behavioral foundation for humanoid tracking and recovery that augments nominal tracking with perturbed get-up references and a hidden successor-state objective, without requiring recovery-specific references or commands at deployment.
  \item Matched G1 experiments covering complete tracking, 100-trial recovery, post-fall motion and load, MoE dynamics, and hardware deployment.
\end{itemize}

\section{Related Work}

Table~\ref{tab:literature_positioning} provides an overview of the task-level distinctions developed in the following discussion.

\subsection{Motion Tracking and Recovery-Enabled Control}
DeepMimic established physics-based reference tracking~\cite{peng2018deepmimic}; AMP broadened skill learning through adversarial motion priors~\cite{peng2021amp}; and BeyondMimic and SONIC scaled humanoid tracking to diverse motions~\cite{liao2025beyondmimic,luo2025sonic}.
KungFuAthlete additionally learns fall-resilient tracking~\cite{lei2026kungfuathlete}, Embrace Collisions targets intentionally contact-rich commands~\cite{zhuang2025embrace}, and BFM-Zero learns broad prompt-conditioned behaviors~\cite{li2025bfmzero}.
StableMimic instead studies an unplanned fall during tracking, reference-shaped recovery without a runtime get-up command, and resumption of the uninterrupted command; quantitative comparisons therefore use only the matched protocol in Sec.~\ref{sec:experiments}.

\subsection{Stand-Up and Fall-Safety Policies}
FRASA studies end-to-end fall recovery~\cite{gaspard2024frasa}, HoST learns stand-up control across postures and terrains~\cite{huang2025host}, and HumanUP separates get-up discovery from deployment-oriented motion refinement~\cite{he2025getup}.
Complementary fall-safety work predicts unavoidable falls and learns protective control (SafeFall)~\cite{meng2025safefall}, optimizes soft terminal poses (Robot Crash Course)~\cite{strauch2025crash}, or unifies prevention, impact mitigation, and recovery from sparse demonstrations (FIRM)~\cite{xu2025firm}.
These methods primarily formulate stand-up or fall safety as a task distinct from continuous command tracking.
Our focus is human-reference-shaped recovery embedded in one always-online motion tracker that subsequently resumes its command.

\subsection{Modular Reinforcement Learning and Asymmetric Training}
Mixture-of-experts routing combines specialized subnetworks~\cite{shazeer2017moe}, while soft modularization reduces destructive sharing in multi-task RL~\cite{yang2020softmod}.
Actor--critic mixtures have specialized locomotion over terrain~\cite{peng2016mace}; later robotic multi-expert policies compose locomotion and recovery skills~\cite{yang2020mela}, and curricula encourage context-dependent specialization~\cite{celik2024diskill}.
CMoE uses terrain-conditioned contrastive routing for multi-terrain humanoid locomotion~\cite{ma2026cmoe}; StableMimic instead uses training-only regime supervision to route between tracking and post-fall recovery.
Unlike a hierarchy invoking separate controllers~\cite{sutton1999options}, StableMimic evaluates both experts continuously and blends their means.
Asymmetric actor--critic training supplies hidden references and simulator state only to value learning and auxiliary supervision~\cite{pinto2017asymmetric}; PPO jointly optimizes the deployable experts and gate~\cite{schulman2017ppo}.

\begin{figure*}[!t]
  \centering
  \includegraphics[width=\textwidth]{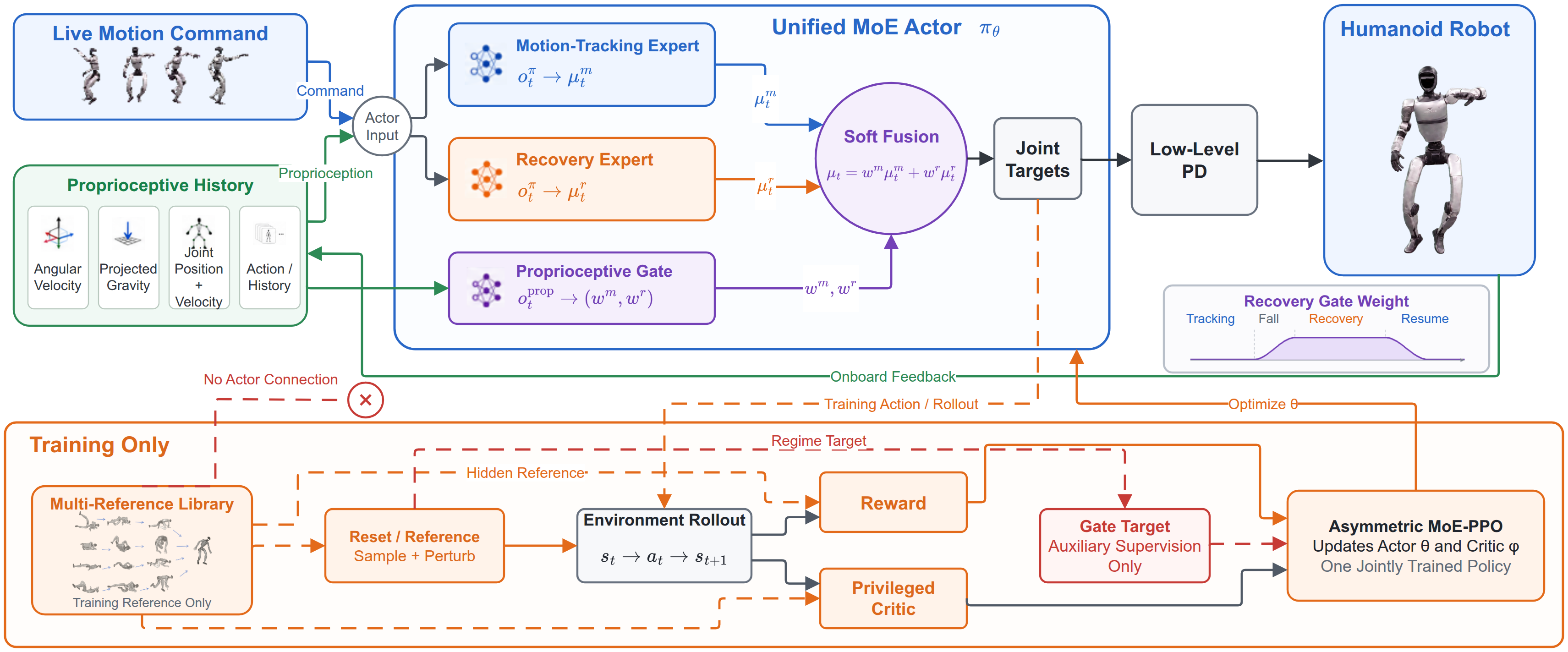}
  \caption{StableMimic method overview. Both experts receive the live command and deployable observation history, while the gate uses proprioceptive history only. The get-up library supplies training-only resets, rewards, and privileged Critic inputs and is removed at deployment. The inset qualitatively illustrates soft recovery routing rather than a measured trace.}
  \label{fig:method_overview}
\end{figure*}

\section{Method}

\subsection{Problem Formulation and Information Boundary}

\begin{figure*}[!t]
  \centering
  \includegraphics[width=\textwidth]{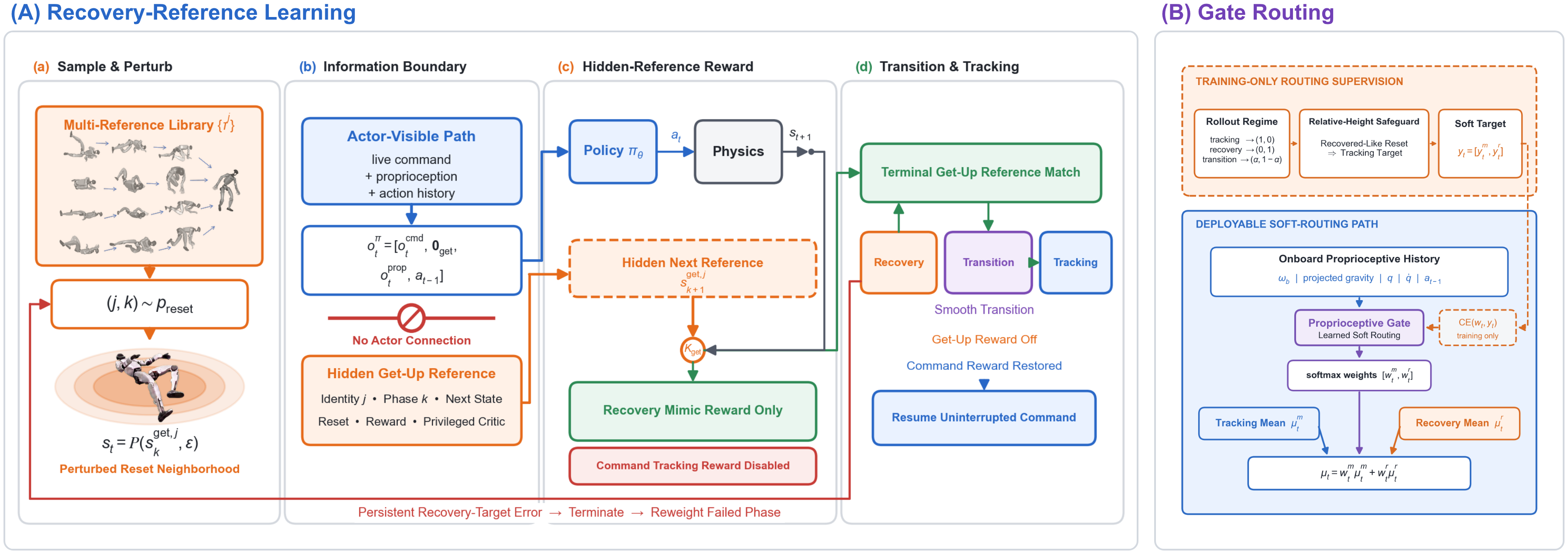}
  \caption{Recovery-reference learning and gate supervision. (A) Perturbed get-up states and a permanently masked reference define the successor-state reward and transition back to command tracking. (B) Rollout labels and the relative-height safeguard form a training-only soft target; deployment retains only proprioceptive routing and continuous expert-mean fusion. Dashed supervision paths are removed from the exported Actor.}
  \label{fig:recovery_gate_training}
  \label{fig:hidden_reference_reward}
  \label{fig:gate_training_logic}
\end{figure*}

StableMimic is one deployable policy with two specialized experts: a commanded-motion expert and a recovery expert, continuously blended by a proprioceptive soft gate (Fig.~\ref{fig:method_overview}).
Both experts receive the live command and deployable observation history, whereas the gate receives proprioception only.
Training-only recovery references construct resets, rewards, and Critic inputs; deployment retains only the jointly trained Actor.
Fig.~\ref{fig:recovery_gate_training} details the recovery objective and gate supervision.

\paragraph{Actor-visible and training-only information}

The deployable observation is
\begin{equation}
  o_t^\pi = \mathcal{H}_4\!\left(o_t^{\mathrm{cmd}},\mathbf{0}_{\mathrm{get}},o_t^{\mathrm{prop}},a_{t-1}\right),
\end{equation}
where $\mathcal{H}_4$ concatenates four consecutive samples.
The live command remains visible throughout falling, recovery, and resumption; $o_t^{\mathrm{prop}}$ contains IMU-compatible signals, joint states, and recent actions.
Fixed-size get-up slots are permanently masked, hence $\mathbf{0}_{\mathrm{get}}$ is zero in every Actor sample.
Training-only get-up features and uncorrupted state enter the Critic; reference ID/frame remain in reset/reward logic, while recovery/transition targets supervise only the Gate loss.
None enters the deployed Actor or serves as a behavior-cloning label (Table~\ref{tab:information_boundary}).

\begin{table}[htbp]
  \centering
  \caption{Training and deployment information boundary.
  \textcolor{blue!65!black}{Training only}: a dagger ($\dagger$) marks quantities removed after optimization;
  \textcolor{blue!65!black}{Gate supervision}: the auxiliary routing target is never provided as a Gate input.}
  \label{tab:information_boundary}
  \resizebox{\columnwidth}{!}{%
  \begin{tabular}{lccccc}
    \toprule
    Information & Experts & Gate & Critic & Reset/reward & Deployment \\
    \midrule
    Live motion command & Yes & No & Yes & Yes & Yes \\
    Proprioceptive history & Yes & Input & Yes & State only & Yes \\
    Get-up reference features & Always zero & No & Active ref.$^\dagger$ & Active ref.$^\dagger$ & Always zero \\
    Reference ID/frame index & No & No & No & Yes$^\dagger$ & No \\
    Recovery/transition target & No & Loss only$^\dagger$ & No & Yes$^\dagger$ & No \\
    Privileged simulator state & No & No & Yes$^\dagger$ & Yes$^\dagger$ & No \\
    \bottomrule
  \end{tabular}}
\end{table}

\subsection{Recovery Support Beyond the Tracking Tube}

\begin{figure}[htbp]
  \centering
  \includegraphics[width=\columnwidth]{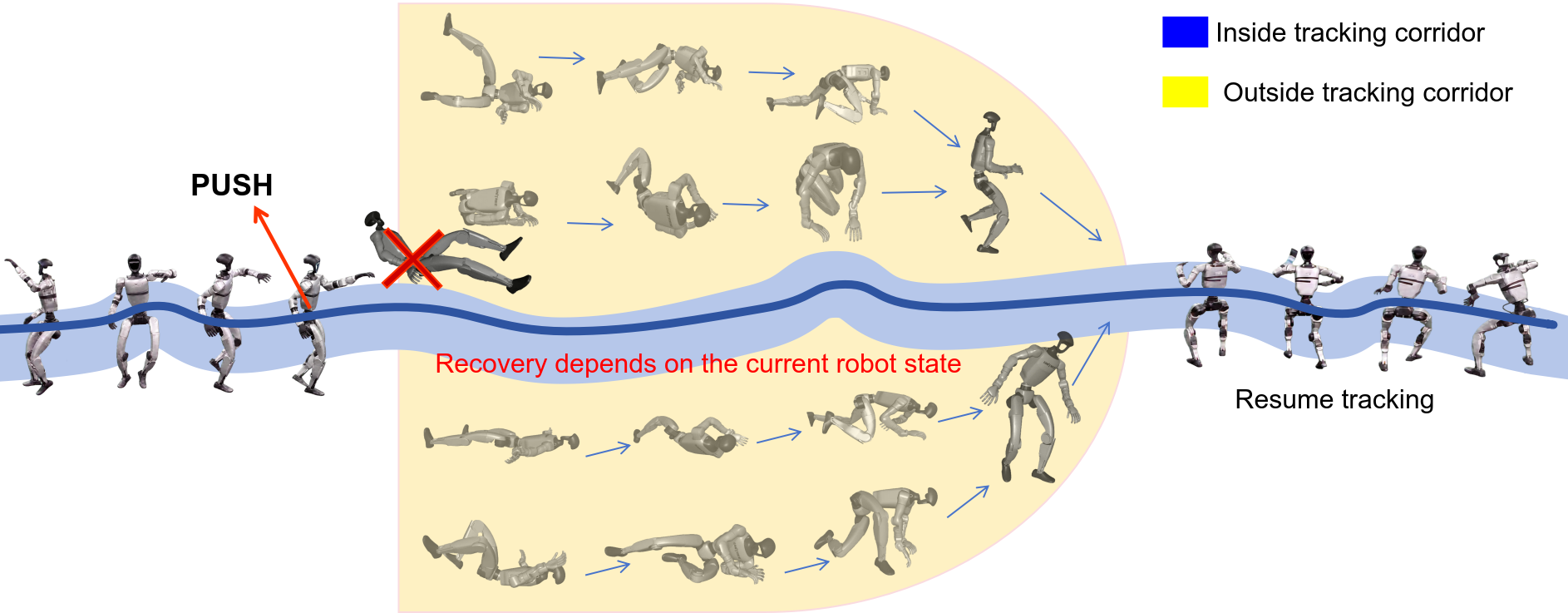}
  \caption{Conceptual state--time view of the learned support. The blue curve and $\mathcal{T}_\epsilon$ denote the command and its locally trackable neighborhood; $\mathcal{R}_\delta$ denotes sampled neighborhoods around perturbed get-up references. A state leaving $\mathcal{T}_\epsilon$ but remaining in $\mathcal{R}_\delta$ is routed toward recovery until tracking becomes feasible again.}
  \label{fig:state_conditioned_recovery}
\end{figure}

Let $\mathcal{T}_\epsilon$ denote a local tube around the commanded state curve and $\mathcal{R}_\delta$ the union of sampled, perturbed neighborhoods around the get-up references (Fig.~\ref{fig:state_conditioned_recovery}).
These are empirical supports, not formal reachability sets.

Training alternates equally between commanded-motion and recovery resets.
For recovery, a trajectory $j$ and frame $k$ are sampled, the corresponding state is perturbed, and the same hidden reference advances for that episode:
\begin{equation}
 \begin{aligned}
 (j,k)&\sim p_{\mathrm{reset}}, &
 s_t&=\mathcal{P}(s_k^{\mathrm{get},j};\epsilon),\\
 a_t&\sim\pi_\theta(o_t), &
 s_{t+1}&\sim p(\cdot\mid s_t,a_t),\\
 r_t^{\mathrm{mimic}}&=K(s_{t+1},s_{k+1}^{\mathrm{get},j}).
 \end{aligned}
 \label{eq:recovery_transition}
\end{equation}
Thus PPO learns actions whose physical successor approaches the next hidden reference state.
Because $j$, $k$, and the reference are masked, this is neither action cloning nor online trajectory retrieval; the policy must infer recovery from its current physical history.

Recovery resets perturb root pose, velocity, and joints around prone, supine, and intermediate reference frames (ranges in Appendix Table~\ref{tab:implementation_details}).
A failure-adaptive sampler mixes hard-frame sampling equally with uniform sampling.
Together, these mechanisms thicken discrete demonstrations into local low-height, contact, and off-balance neighborhoods; states beyond this sampled support are outside our claim.

During recovery, termination and reward are evaluated against the active get-up target, not the still-visible command.
Target error persisting for $2.0\,\mathrm{s}$ terminates the episode, while terminal similarity initiates a $1.5$-s transition back to the advancing command.
Failed phase bins receive greater future sampling probability.

Fig.~\ref{fig:state_conditioned_recovery} summarizes the relationship between the local command-tracking tube and the sampled recovery support.

\subsection{Proprioception-Gated Dual-Expert Actor}

The Single-MLP baseline routes tracking and recovery advantages through one parameter set, allowing updates from either regime to perturb features needed by the other.
StableMimic instead applies soft modularization~\cite{yang2020softmod} while retaining a shared observation/action interface.

Each expert maps the same command-conditioned deployable history to a joint-target mean, while the gate uses only the proprioceptive part of that history: base angular velocity, projected gravity, joint position, joint velocity, and previous action.  It contains neither command/reference features nor a recovery-stage variable.  Routing and action blending are
\begin{equation}
 \begin{aligned}
 \mu_t^m&=f_m(o_t^\pi),& \mu_t^r&=f_r(o_t^\pi),\\
 [w_t^m,w_t^r]&=\operatorname{softmax}(f_g(o_t^{\mathrm{prop}})/\tau),\\
 \mu_t&=w_t^m\mu_t^m+w_t^r\mu_t^r.
 \end{aligned}
 \label{eq:moe_policy}
\end{equation}
The policy samples a Gaussian with shared learned variance around $\mu_t$; soft blending avoids a discrete action switch.

\paragraph{Auxiliary gate supervision}

During training only, the rollout regime defines the gate target $y_t$: tracking favors $(1,0)$, recovery favors $(0,1)$, and the $1.5$-s transition interpolates smoothly.
A recovered-like reset retains the tracking target when $h_{\mathrm{base}}^{\mathrm{reset}}\geq0.8h_{\mathrm{cmd}}$.
Cross-entropy promotes specialization, within-regime temporal consistency regularizes routing, and transition samples align expert outputs.
These labels and the relative-height safeguard are privileged supervision, never policy inputs or runtime switches (Fig.~\ref{fig:recovery_gate_training}(B)).

Command-near states, including low or ground-contact poses, always retain tracking targets; consequently, the four-frame Gate learns recovery-specific temporal proprioceptive patterns rather than treating any commanded pose as a recovery cue.

\paragraph{Recovery objective}

Whole-body imitation shapes recovery toward the demonstrated human motion, while action-rate, joint-limit, self-collision, and power penalties discourage aggressive corrections; these choices are load-aware but do not certify safety.

Command and get-up targets use the same Gaussian-kernel form and the same six error families: root position/orientation and body position/orientation/linear-velocity/angular-velocity.  Command tracking uses the full aligned root-position error, whereas recovery uses root height so that a translated get-up remains valid.  Let $K_{\mathrm{cmd}}$ and $K_{\mathrm{get}}$ denote the two unscaled similarity kernels, and let $\lambda_{\mathrm{cmd}}$, $\lambda_{\mathrm{get}}$, and $\lambda_{\mathrm{trans}}$ denote the regime coefficients.  The reference-tracking part of the reward is selected by the training regime:
\begin{equation}
 \begin{aligned}
 K_{c_t}(s,\bar s)&=\sum_{i=1}^{6}w_i
 \exp\!\left(-\|e_i^{c_t}(s,\bar s)\|^2/\sigma_i^2\right),\\
 r_t^{\mathrm{ref}}&=
 \begin{cases}
 \lambda_{\mathrm{get}}K_{\mathrm{get}}(s_{t+1},s_{k+1}^{\mathrm{get},j}), & \text{recovery},\\
 \alpha_t\lambda_{\mathrm{trans}}K_{\mathrm{cmd}}(s_{t+1},s_{t+1}^{\mathrm{cmd}}), & \text{transition},\\
 \lambda_{\mathrm{cmd}}K_{\mathrm{cmd}}(s_{t+1},s_{t+1}^{\mathrm{cmd}}), & \text{tracking},
 \end{cases}\\
 r_t&=r_t^{\mathrm{ref}}+r_t^{\mathrm{reg}}+r_t^{\mathrm{success}}.
 \end{aligned}
 \label{eq:reward}
\end{equation}
During recovery, the live command remains observable but its tracking reward is zero; only the hidden get-up target supplies the reference term.
Its coefficient is $2.5$ times the nominal command coefficient, which strengthens this phase without implying a $2.5\times$ realized return.
Regularization remains active, and the transition removes the get-up term while introducing command tracking smoothly.
The demonstrated style therefore shapes behavior only within the sampled perturbation support.

\subsection{Optimization and Get-Up-Reference-Free Deployment}

We optimize the Actor and privileged Critic with standard clipped PPO~\cite{schulman2017ppo}.
The loss additionally contains gate-target cross-entropy, within-regime routing consistency, and transition expert-alignment terms; gradients through Eq.~\eqref{eq:moe_policy} route most tracking and recovery updates toward their corresponding experts.
Failed recovery truncates return, whereas successful recovery retains subsequent command-tracking return.

The Single-MLP ablation shares the observation, action, Critic, and PPO settings but uses a widened 1024--512--256 backbone, with twice the hidden width of each expert at every layer.
Its 1.57-M-parameter Actor is larger than the 1.30-M-parameter MoE Actor, so increased total Actor capacity alone cannot explain an MoE advantage.
Dimensions and optimization settings appear in Appendix Table~\ref{tab:implementation_details}.

\paragraph{Get-up-reference-free deployment and tracking resumption}

At deployment, the library, phase labels, rewards, and Critic are absent; there is no get-up-reference lookup, recovery command, or hard switch.
The gate infers routing from proprioceptive history and returns authority to tracking after command reacquisition.

\paragraph{Reference alignment}

Horizontal translation is re-aligned to the recovered pose while command yaw and local motion continuity are retained, avoiding pursuit of an unreachable pre-fall position.

\paragraph{Training setup}

Training uses equal tracking/recovery episode probability, observation corruption, and dynamics randomization; Appendix Table~\ref{tab:implementation_details} collects numerical settings.

\section{Experiments}
\label{sec:experiments}

\subsection{Experimental Setup}

\paragraph{Tasks and protocols}
Experiments use the Unitree G1 and retargeted LAFAN1 motions~\cite{harvey2020lafan}.
The training-only get-up library is constructed from the retargeted LAFAN1 get-up subset; tracking commands use its dance subset, and the two are sequence-disjoint.
We evaluate two coupled capabilities: complete-sequence tracking over all LAFAN1 dance sequences, and post-fall recovery followed by command resumption.
All policies run in one nominal MuJoCo G1 environment with common actuator limits and metrics, a $0.005\,\mathrm{s}$ simulation step, and deterministic actions at $50\,\mathrm{Hz}$.
Tracking runs to sequence completion, whereas recovery uses a $20\,\mathrm{s}$ horizon.
Test randomization, observation corruption, delay, and early termination are disabled.

The recovery protocol uses the same 100 pre-generated disturbances for every method: 25 per direction in $\{+x,-x,+y,-y\}$, each a $0.2\,\mathrm{s}$ horizontal torso force stratified from $525$ to $575\,\mathrm{N}$.
Pose-conditioned policies receive a held command, KungFuAthlete an advancing command, and BFM-Zero its native goal interface.
StableMimic retains its live command and proprioception but receives no get-up reference, phase, motion ID, recovery flag, or target action.

\paragraph{Comparison methods}
We compare StableMimic with its Single-MLP ablation, a BeyondMimic-style tracker~\cite{liao2025beyondmimic}, KungFuAthlete~\cite{lei2026kungfuathlete}, and BFM-Zero~\cite{li2025bfmzero}.
The Single-MLP policy shares the observation, action, Critic, reward, and PPO settings of StableMimic but replaces the gated experts with one widened 1024--512--256 backbone whose parameter count exceeds that of the complete MoE Actor.
BeyondMimic provides a tracking-only comparison, while KungFuAthlete and BFM-Zero provide recovery-capable comparisons through their supported command interfaces.
All baselines retain their original training procedures.

\paragraph{Evaluation metrics}
For tracking, we exclude world translation and heading drift.
For each of the $B=14$ common key bodies, measured and reference positions are expressed relative to their respective pelvis and rotated into their respective pelvis-yaw frames, yielding $\bar p_{t,b}$ and $\bar p^{\mathrm{ref}}_{t,b}$.
Our primary metric is
\begin{equation}
  E_{\mathrm{MPBPE}} = \frac{10^3}{TB}\sum_{t,b}
  \lVert\bar p_{t,b}-\bar p^{\mathrm{ref}}_{t,b}\rVert_2 .
\end{equation}
We additionally report mean absolute joint-angle error (MJAE), joint-velocity error (MJAVE), and the Cartesian second-difference error of the same pelvis-relative bodies.

A fall begins when pelvis height drops below $0.50\,\mathrm{m}$ or pelvis tilt exceeds $60^\circ$.
Recovery requires a structured get-up and tracking resumption without resetting the policy or command, and all 100 trials remain in the success denominator.
Motion, control-output, and load metrics are computed over the fixed three seconds after the first fallen frame across all trials.
Instantaneous motion, target-rate, and simulated actuator-torque signals use the within-trial 95th percentile, whereas limb path and positive energy are window integrals.
Here $q^{\mathrm{tar}}$ is the physical PD joint-position target after action scaling, and $\tau$ is the executed simulated actuator torque after controller and effort limiting.
Pelvis displacement spans fall to detected recovery, or rollout end if unsuccessful.

\subsection{Main Simulation Results}

\subsubsection{Complete-Sequence Motion Tracking}

All methods traverse the complete LAFAN1 dance stream from beginning to end.
A common initialization interval is executed but excluded from aggregation.

\begin{table}[htbp]
  \centering
  \caption{Complete-sequence tracking over all retargeted LAFAN1 dance sequences.
  \textcolor{blue!65!black}{Protocol}: all methods use the same deterministic evaluation;
  \textcolor{blue!65!black}{Alignment}: body-position errors are root-yaw aligned;
  \textcolor{blue!65!black}{Ranking}: lower is better, and boldface marks the best result.}
  \label{tab:tracking_results}
  \scriptsize
  \setlength{\tabcolsep}{1.8pt}
  \renewcommand{\arraystretch}{1.08}
  \begin{tabular*}{\columnwidth}{@{\extracolsep{\fill}}lrrrr@{}}
    \toprule
    Method &
    \shortstack{MPBPE\\(mm)} &
    \shortstack{MJAE\\($10^{-3}$ rad)} &
    \shortstack{MJAVE\\($10^{-3}$ rad/s)} &
    \shortstack{Rel. acc.\\(mm/frame$^2$)} \\
    \midrule
BeyondMimic & 32.37 & 104.50 & 893.41 & 2.29 \\
KungFuAthlete & 56.04 & 162.23 & 1156.36 & 2.65 \\
BFM-Zero & 249.40 & 444.28 & 2309.79 & 9.44 \\
StableMimic (Single MLP) & 32.66 & 113.56 & 846.54 & 2.09 \\
StableMimic (MoE) & \textbf{28.53} & \textbf{88.83} & \textbf{798.22} & \textbf{2.08} \\
\bottomrule

  \end{tabular*}
\end{table}

\paragraph{Overall tracking performance}
Table~\ref{tab:tracking_results} and Fig.~\ref{fig:quantitative_summary}(a)--(d) show that StableMimic attains the lowest error in all four reported tracking metrics.
It achieves an MPBPE of $\TrackingMoeMPBPE\,\mathrm{mm}$ and an MJAE of $\TrackingMoeMJAE\times10^{-3}\,\mathrm{rad}$ while retaining the same recovery-capable policy interface.
The result demonstrates that extending the state support beyond nominal tracking does not require sacrificing complete-sequence tracking accuracy.

\paragraph{Effect of expert specialization}
Relative to the Single-MLP policy, StableMimic reduces MPBPE from $\TrackingMlpMPBPE$ to $\TrackingMoeMPBPE\,\mathrm{mm}$ and MJAE from $\TrackingMlpMJAE$ to $\TrackingMoeMJAE\times10^{-3}\,\mathrm{rad}$; it also lowers MJAVE and relative body-acceleration error.
This consistent improvement despite the wider, higher-capacity Single-MLP is compatible with the intended separation of tracking and recovery updates and cannot be attributed simply to a larger MoE parameter count.

\subsubsection{Post-Fall Recovery}

\begin{figure}[htbp]
  \centering
  \includegraphics[width=\columnwidth]{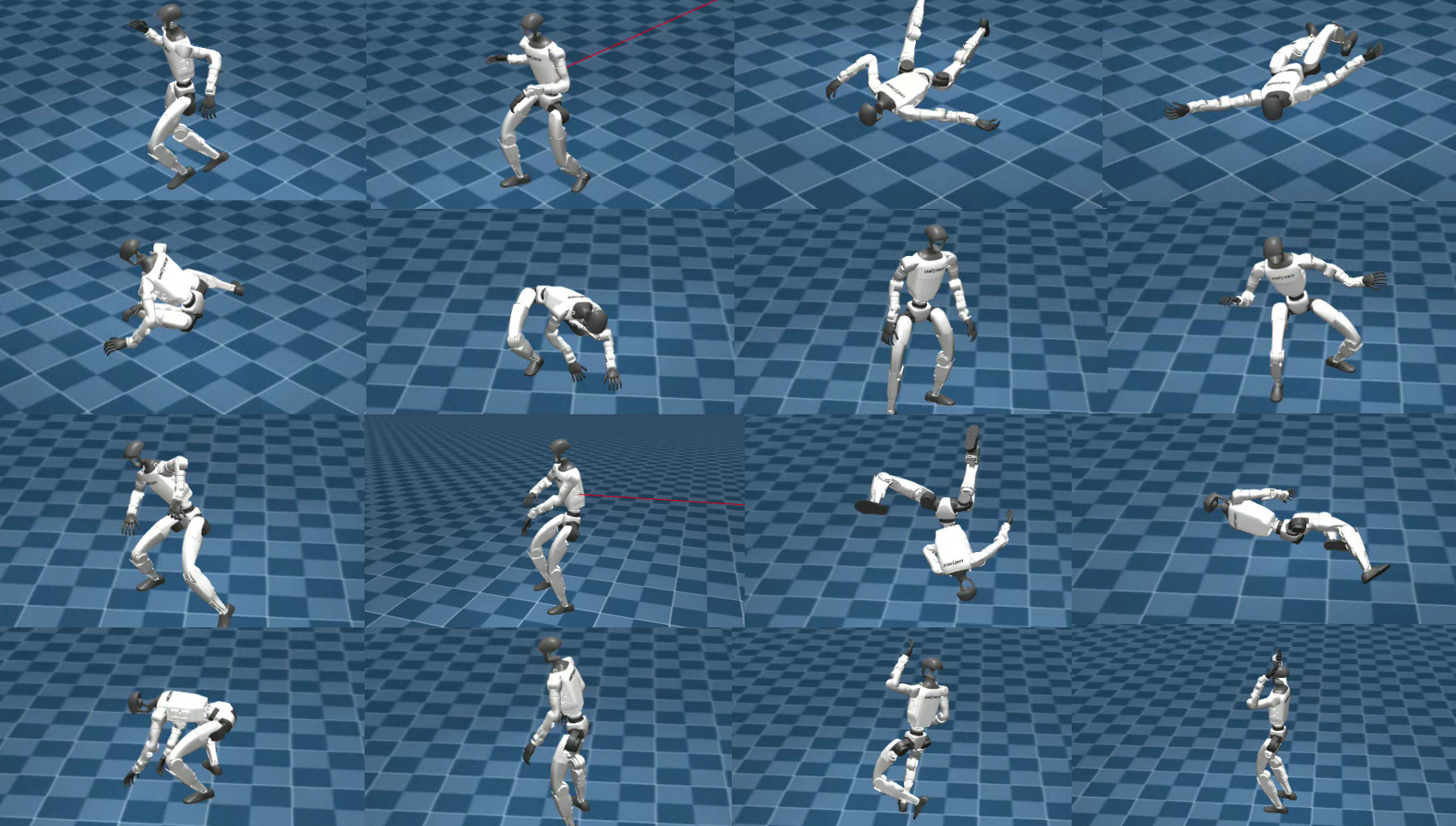}
  \caption{Representative tracking--fall--recovery rollouts in MuJoCo. Top: prone fall and command reacquisition. Bottom: supine fall and command reacquisition. The live command remains active, while no get-up command, reference phase, or motion identifier is provided.}
  \label{fig:mujoco_recovery_sequences}
\end{figure}

Figure~\ref{fig:mujoco_recovery_sequences} shows representative prone and supine recovery sequences under the unchanged deployment interface.
The live command remains active throughout the fall, recovery, and command-reacquisition phases.

\begin{center}
  \begin{minipage}{\columnwidth}
    \centering
    \includegraphics[width=\linewidth]{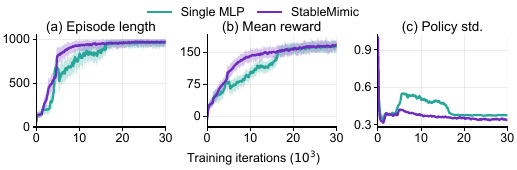}
    \captionof{figure}{Training dynamics over $30{,}000$ iterations. StableMimic stabilizes episode length and reward earlier and reaches a lower policy standard deviation than Single-MLP.}
    \label{fig:training_curves}
  \end{minipage}
\end{center}

\begin{figure*}[!t]
  \centering
  \includegraphics[width=0.98\textwidth]{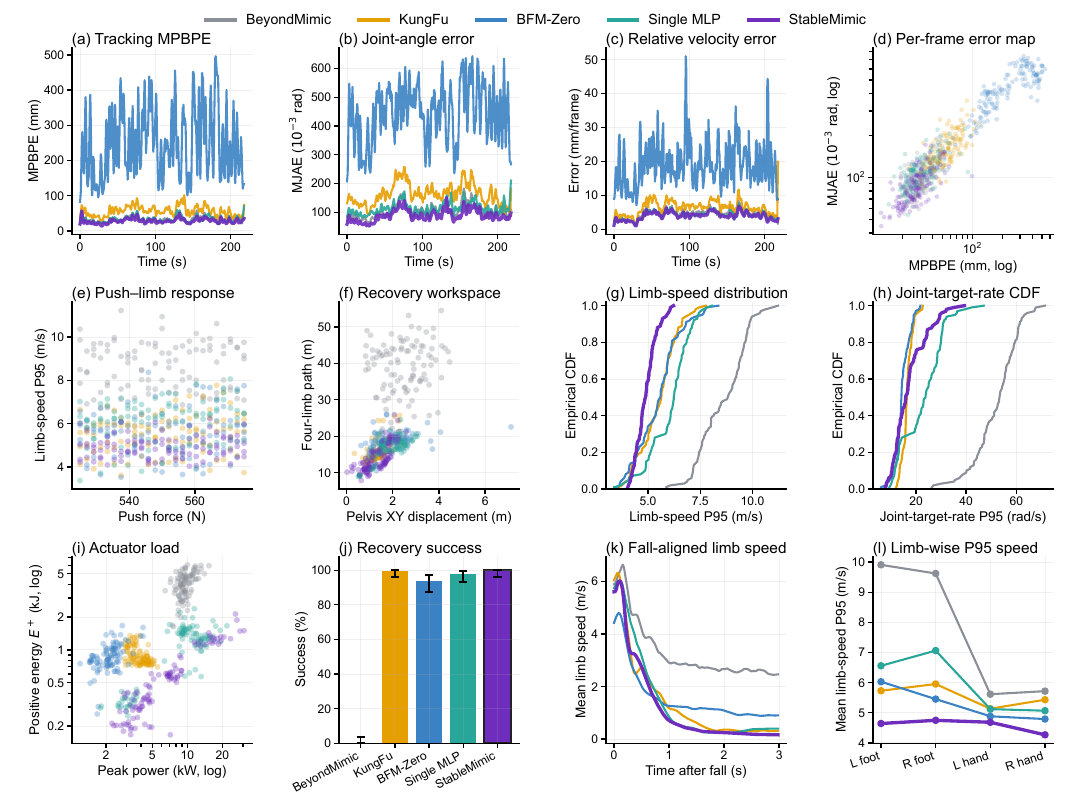}
  \caption{Tracking and post-fall evaluation on the complete dance subset of LAFAN1. (a)--(c) Temporal tracking errors; (d) per-sample body-position versus joint-angle error; (e)--(f) disturbance response and recovery workspace; (g)--(i) post-fall motion and actuator-load distributions; and (j)--(l) recovery success, fall-aligned limb speed, and limb-wise P95 speed. All recovery panels use the same 100 disturbances per method; points denote individual trials or temporal samples.}
  \label{fig:quantitative_summary}
\end{figure*}

\begin{table*}[!b]
  \centering
  \caption{Post-fall behavior over 100 disturbances.
  \textcolor{blue!65!black}{Window}: motion, control-output, and load entries average per-trial statistics over three seconds;
  \textcolor{blue!65!black}{Percentiles}: superscript 95 denotes a within-trial percentile;
  \textcolor{blue!65!black}{Integrals}: $L_{\mathrm{limb}}$ and $E^+$ accumulate over the same window;
  \textcolor{blue!65!black}{Workspace}: $\Delta p_{xy}$ spans fall to recovery, or rollout end if unsuccessful.
  Lower is better except for success; boldface marks the best result.}
  \label{tab:push_fall_safety}
  \scriptsize
  \setlength{\tabcolsep}{2.0pt}
  \renewcommand{\arraystretch}{1.08}
  \begin{tabular*}{\textwidth}{@{\extracolsep{\fill}}lrrrrrrrr@{}}
    \toprule
    & Outcome & \multicolumn{3}{c}{Bounded post-fall motion} & Control output & Workspace & \multicolumn{2}{c}{Actuator load} \\
    \cmidrule(lr){2-2}\cmidrule(lr){3-5}\cmidrule(lr){6-6}\cmidrule(lr){7-7}\cmidrule(lr){8-9}
    Method &
    \shortstack{Succ.\\$\uparrow$} &
    \shortstack{$v_{\mathrm{limb}}^{95}$\\(m/s) $\downarrow$} &
    \shortstack{$L_{\mathrm{limb}}$\\(m) $\downarrow$} &
    \shortstack{$|\dot q|^{95}$\\(rad/s) $\downarrow$} &
    \shortstack{$|\Delta q^{\mathrm{tar}}/\Delta t|^{95}$\\(rad/s) $\downarrow$} &
    \shortstack{$\Delta p_{xy}$\\(m) $\downarrow$} &
    \shortstack{$|\tau|^{95}$\\(N\,m) $\downarrow$} &
    \shortstack{$E^+$\\(kJ) $\downarrow$} \\
    \midrule
BeyondMimic & 0/100 & 8.62 & 38.96 & 14.38 & 52.08 & 2.48 & 63.74 & 4.10 \\
KungFuAthlete & \textbf{100/100} & 5.51 & 16.18 & 6.49 & 16.47 & 1.44 & 37.42 & 0.90 \\
BFM-Zero & 94/100 & 5.51 & 18.42 & 6.22 & \textbf{14.60} & 1.90 & 31.87 & 0.89 \\
StableMimic (Single MLP) & 98/100 & 6.06 & 16.04 & 6.34 & 22.80 & 1.78 & 27.94 & 1.10 \\
StableMimic (MoE) & \textbf{100/100} & \textbf{4.91} & \textbf{13.80} & \textbf{5.26} & 17.35 & \textbf{1.32} & \textbf{19.64} & \textbf{0.61} \\
\bottomrule

  \end{tabular*}
\end{table*}

\paragraph{Overall post-fall performance}
As reported in Table~\ref{tab:push_fall_safety} and Fig.~\ref{fig:quantitative_summary}(e)--(l), StableMimic recovers in $\RecoveryMoeSuccess$ disturbances and obtains the lowest limb-speed P95, limb travel, joint-speed P95, recovery displacement, actuator-torque P95, and positive energy among the evaluated methods.
These complementary measurements characterize bounded post-fall motion without collapsing smoothness into one composite score.
Lower limb speed and swept distance reduce potential impact and intrusion into nearby space; together with lower joint motion and actuator load, these results provide direct empirical evidence of improved post-fall interaction safety.

\paragraph{Comparison with tracking-only control}
BeyondMimic does not recover under this protocol and exhibits the largest limb motion, joint speed, target rate, recovery displacement, actuator torque, and positive energy.
This behavior is consistent with the structural difficulty of continuing to minimize tracking error when the live command is temporarily unreachable.
StableMimic instead routes such states through human-reference-shaped recovery before reacquiring that command.

\paragraph{Comparison with recovery-capable policies}
KungFuAthlete also recovers in all 100 trials, but StableMimic requires less limb travel, joint motion, recovery workspace, actuator torque, and positive energy.
BFM-Zero attains the lowest target rate, yet StableMimic achieves a higher recovery rate and lower values for the remaining reported motion and load measures.

\paragraph{Effect of expert specialization}
The Single-MLP and StableMimic policies recover in $\RecoveryMlpSuccess$ and $\RecoveryMoeSuccess$ trials, respectively.
StableMimic reduces limb-speed P95, limb travel, joint-speed P95, actuator-torque P95, and positive energy from $\RecoveryMlpLimbSpeedTail$, $\RecoveryMlpLimbPath$, $\RecoveryMlpJointSpeed$, $\RecoveryMlpTorqueTail$, and $\RecoveryMlpEnergy$ to $\RecoveryMoeLimbSpeedTail\,\mathrm{m/s}$, $\RecoveryMoeLimbPath\,\mathrm{m}$, $\RecoveryMoeJointSpeed\,\mathrm{rad/s}$, $\RecoveryMoeTorqueTail\,\mathrm{N\,m}$, and $\RecoveryMoeEnergy\,\mathrm{kJ}$, respectively.

\begin{figure*}[!t]
  \centering
  \includegraphics[width=0.88\textwidth]{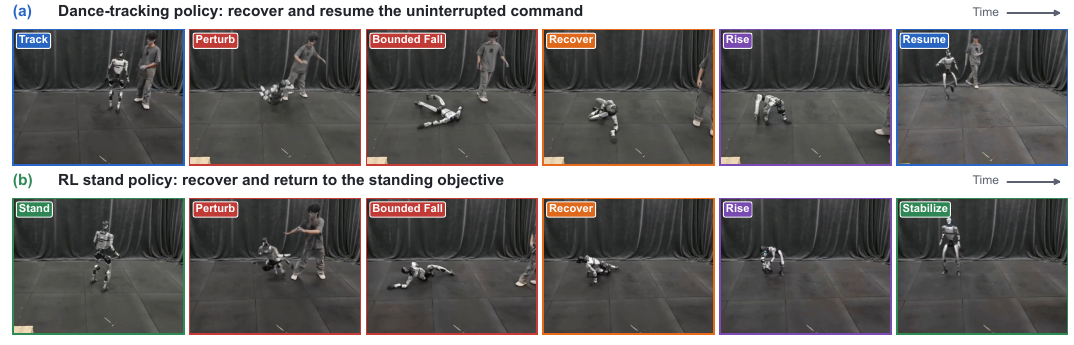}
  \caption{Real-G1 deployment of dance (top) and standing-reference (bottom) policies, showing recovery and task resumption with bounded limb motion.}
  \label{fig:real_robot_deployment}
\end{figure*}

\subsection{Analysis of the MoE Design}

Figure~\ref{fig:training_curves} provides descriptive optimization evidence complementary to the final-policy metrics.
Compared with Single-MLP, StableMimic stabilizes episode length and reward earlier and reaches a lower policy standard deviation.
Because both policies use the same entropy coefficient, this behavior is consistent with reduced interference under expert routing, although the curves alone remain descriptive rather than causal evidence of expert independence.

\subsection{Main Hardware Results}

Each Actor is exported as one ONNX policy that handles tracking, fall, recovery, and command resumption, and runs on the G1 at $50\,\mathrm{Hz}$ with the training observation order, four-frame history, joint mapping, action scale, and position-control interface.
Only the live command and onboard proprioception are supplied; hardware joint, velocity, torque, timeout, and emergency-stop safeguards remain active.
We deploy a dance tracker and, as a lightweight extension, a separately trained instance whose command is a constant standing joint-angle reference.
Figure~\ref{fig:real_robot_deployment} summarizes the two qualitative deployments.

\paragraph{Dance tracking}
After a human-induced fall, the dance policy follows a structured recovery path and resumes its uninterrupted command without an external trigger or policy restart.

\paragraph{Static-standing extension}
Replacing the dance trajectory with a constant standing joint-angle reference yields an RL-stand tracker that recovers and restabilizes after perturbation, illustrating a possible extension toward more bounded post-disturbance behavior that is more benign to the robot and its surroundings.

\paragraph{Scope}
In both sequences, the fallen robot exhibits bounded limb motion and returns to a feasible task region rather than persistently chasing the unavailable target.
This is qualitative evidence of physical-robot transfer, not a safety certification or a matched hardware architecture comparison.

\section{Conclusion}

StableMimic integrates whole-body command tracking and human-reference-shaped post-fall recovery in a single proprioception-gated dual-expert policy.
Perturbed resets around multiple get-up references enlarge the training support beyond the nominal tracking tube, while a hidden successor-state objective teaches recovery without exposing the Actor to trajectory identity, recovery phase, or a runtime get-up command.
The soft gate separates tracking- and recovery-dominant updates, routes fallen states toward structured recovery behavior, and returns authority to tracking after command reacquisition.
Across the matched G1 evaluation, the MoE preserves complete-sequence tracking while reducing several post-fall motion and actuator-load measures relative to Single-MLP.
Real-robot dance and constant-standing-reference policies further demonstrate bounded limb motion, autonomous recovery, and task resumption under the same deployment interface.
By defining structured behavior outside the command distribution, the proposed recovery-support curriculum improves post-fall interaction safety under the evaluated protocol and can augment teleoperation-oriented and other RL trackers with compatible observation and action spaces, although it is not a universal safety guarantee.

\appendices
\section{Implementation Details}

\begin{center}
\vspace{-0.15em}
\captionof{table}{StableMimic and Single-MLP implementation details.}
  \label{tab:implementation_details}
  \fontsize{6.3}{5.8}\selectfont
  \setlength{\tabcolsep}{0.5pt}
  \setlength{\aboverulesep}{0.15ex}
  \setlength{\belowrulesep}{0.15ex}
  \renewcommand{\arraystretch}{0.68}
  \begin{tabular}{@{}p{0.10\columnwidth}p{0.85\columnwidth}@{}}
    \toprule
    Net
      & Inputs (Actor/gate/Critic): $4{\times}221=884$/372/1428; experts 512--256--128, Single MLP 1024--512--256, ELU, 29-D mean; shared scalar standard deviation; Actor/Critic normalization; parameters (MoE/Single/Critic): 1,298,301/1,569,850/896,001. \\
    \midrule
    PPO
      & $\gamma=0.99$; GAE $\lambda=0.95$; clip 0.2; adaptive learning rate 0.001; 24 steps, 5 epochs, 4 mini-batches; gradient norm $\leq1.0$; gate weights: CE 0.1, transition 4.0, consistency 0.01, alignment 0.01; entropy 0.05 for both Actors. \\
    \midrule
    Train
      & 4096 environments; 50-Hz control; 20-s horizon; equal tracking/recovery sampling with failure-adaptive frames; 1.5-s transition; 2.0-s error tolerance; recovery coefficient $2.5{\times}$ nominal. \\
    \midrule
    Reset
      & Position: $x,y\in[-0.05,0.05]$ m, $z\in[-0.01,0.01]$ m; roll/pitch $\in[-0.1,0.1]$ rad; yaw $\in[-0.2,0.2]$ rad.
        Velocity: linear $x,y\in[-0.5,0.5]$ m/s, $z\in[-0.2,0.2]$ m/s; angular roll/pitch $\in[-0.52,0.52]$ rad/s, yaw $\in[-0.78,0.78]$ rad/s.
        Joint noise $[-0.1,0.1]$ rad where enabled. \\
    \bottomrule
  \end{tabular}
\vspace{-1.0em}
\end{center}

\end{document}